# Evaluating model calibration in classification

**Juozas Vaicenavicius**
Uppsala University; Veoneer Inc.

**David Widmann**
Uppsala University

**Carl Andersson**
Uppsala University

**Fredrik Lindsten**
Linköping University

**Jacob Roll**
Veoneer Inc.

**Thomas B. Schön**
Uppsala University

## Abstract

Probabilistic classifiers output a probability distribution on target classes rather than just a class prediction. Besides providing a clear separation of prediction and decision making, the main advantage of probabilistic models is their ability to represent uncertainty about predictions. In safety-critical applications, it is pivotal for a model to possess an adequate sense of uncertainty, which for probabilistic classifiers translates into outputting probability distributions that are consistent with the empirical frequencies observed from realized outcomes. A classifier with such a property is called *calibrated*. In this work, we develop a general theoretical calibration evaluation framework grounded in probability theory, and point out subtleties present in model calibration evaluation that lead to refined interpretations of existing evaluation techniques. Lastly, we propose new ways to quantify and visualize miscalibration in probabilistic classification, including novel multidimensional reliability diagrams.

## 1 Introduction

Understanding whether the predictions of a classification model are trustworthy is of crucial importance in machine learning applications. Many popular classifiers such as neural networks output a real-valued vector whose values are typically used to choose the most likely class or even rank the likeliness of different classes. Hence, the key question in the evaluation of such classifiers is whether the classifier output can be interpreted as real-world probabilities, which are what matters for decision making in reality. As much of machine learning research and applications concern building models with good predictive performance, the question of how well the confidence score (expressed as a number between 0 and 1) of the predicted class is calibrated has been dominating the model evaluation literature (Niculescu-Mizil and Caruana 2005; Guo et al. 2017; Kumar, Sarawagi, and Jain 2018). Put differently, it is whether the confidence score of the predicted class can be interpreted as the probability of the classifier getting the class right–a natural question in many applications.

However, in a number of new, especially safety-critical, applications of machine learning it is of increasing importance to know whether the entire classifier output can be interpreted probabilistically, not just the confidence of the predicted class; this is the main question to be addressed in this paper. We illustrate the need for a probabilistic interpretation of the entire classifier output via a simplistic yet illustrative example. Suppose we have a classification problem in which given an image containing either a single or no living entity a classifier outputs a probability vector expressing the likeliness of three different classes "`no creature`", "`person`", and "`animal`". Suppose that for the same image the outputs of two different classifiers are (`no creature` = 0.9, `person` = 0.1, `animal` = 0) and (`no creature` = 0.9, `person` = 0, `animal` = 0.1). Since all widely adopted calibration evaluation techniques take into account only the score of the predicted class, the two mentioned classifier outputs are the same, as far as evaluation is concerned. However, if these outputs actually corresponded to real-world probabilities, the control actions based on these vectors might be very different, e.g., for an autonomous vehicle, due to behavioral differences between pedestrians and animals or differing driving norms in proximity to objects of the distinct classes. This simple example motivates the significance of our quest for calibration evaluation of probabilistic multiclass classifiers beyond just the confidence score of the predicted class.

In this paper, we develop and motivate a general mathematical framework grounded in probability theory for evaluating probabilistic multiclass classifiers. Within it, the existing cal-ibration evaluation techniques can be seen as special cases. The theoretical construct allows for rigorous in-depth analysis of chosen aspects of model calibration and provides clear statistical interpretation of results. Moreover, it unlocks new and sheds light on existing calibration evaluation methods. In particular, we build up theoretical understanding of how popular calibration evaluation techniques should be reinterpreted to avoid jumping to attractive though unjustified conclusions with possibly undesirable practical consequences. Furthermore, drawing inspiration from the statistics literature we revisit classical reliability diagrams used for calibration evaluation and put forward our variant that conveniently summarizes the relevant calibration information in an interpretable way. Addressing the multidimensional nature of calibration evaluation in a multiclass setting we propose novel multidimensional reliability diagrams that are capable of representing the full calibration evaluation information for classification problems with three and four classes. The proposed calibration evaluation methods are illustrated on standard neural network classifiers.[1]

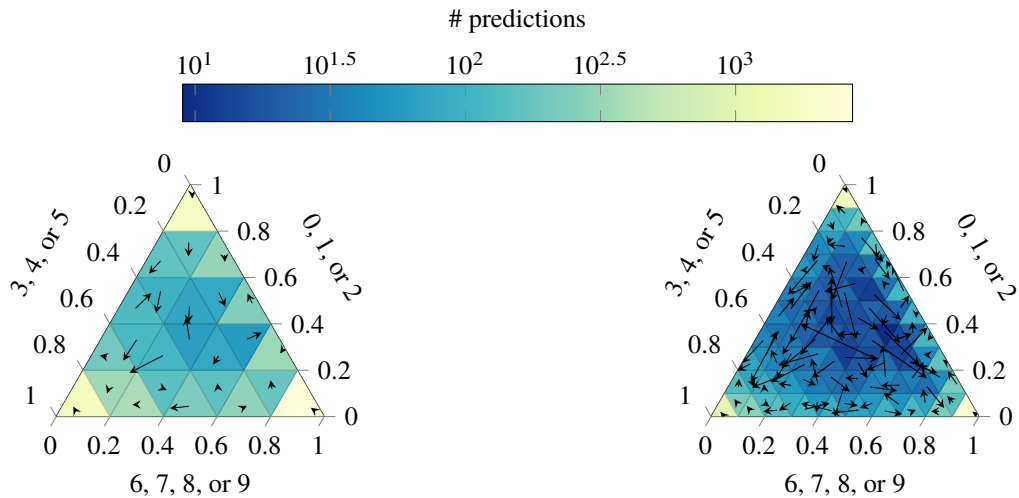

Figure 1: Two-dimensional reliability diagrams for LeNet on the CIFAR-10 test set with 25 and 100 bins of equal size. The predictions are grouped into three groups $\{0, 1, 2\}$, $\{3, 4, 5\}$, and $\{6, 7, 8, 9\}$ of the original classes. Arrows represent the deviation of the estimated calibration function value (arrow head) from the group prediction average (arrow tail) in a bin. The empirical distribution of predictions is visualized by color-coding the bins.

In the NeurIPS 2017 spotlight paper by Lakshminarayanan, Pritzel, and Blundell (2017), the authors conclude with "We hope that our work will encourage the community to consider non-Bayesian approaches (such as ensembles) and other interesting evaluation metrics for predictive uncertainty". Sharing the same interest in predictive uncertainty evaluation, we respond to the call with this article.

## 2 Problem formulation

In this paper, we study a classical supervised classification problem. Let $(\Omega, \mathscr{F}, \mathbb{P})$ be a probability space on which we have independent and identically distributed random variable pairs $(X, Y), (X_1, Y_1), \ldots, (X_n, Y_n) \in \mathscr{X} \times \mathscr{Y}$ having the same joint distribution $p(X, Y)$. Here $\mathscr{X}$ denotes the input space while $\mathscr{Y}$ denotes the finite set of $m$ available class labels. Given a training data set of realizations $\{(x_i, y_i)\}_{i=1}^n$ of $\{(X_i, Y_i)\}_{i=1}^n$, ideally we would like to find the optimal probabilistic classifier $g^{\text{opt}} := p(Y|X = \cdot)$; here the popular "·" notation acts as a placeholder for the argument of a function.

Unfortunately, it is generally impossible to recover the optimal classifier given only a data sample of finite size. Acknowledging this inexorable limitation we can only hope to find a measurable model $g \colon \mathscr{X} \to \mathscr{P}(\mathscr{Y})$ with $g \approx g^{\text{opt}}$ that best suits our purpose. Here $\mathscr{P}(\mathscr{Y})$ denotes the space of probability measures on $\mathscr{Y}$. Since $\mathscr{Y}$ is a finite discrete set, we can identify the space $\mathscr{P}(\mathscr{Y})$ with the $(m - 1)$-dimensional probability simplex $\Delta_{m-1} := \{x \in [0, 1]^m \colon \|x\|_1 = 1\}$. As it is often more convenient to work with vectors in $\Delta_{m-1}$ than with elements in $\mathscr{P}(\mathscr{Y})$, we will use both spaces interchangeably.

In safety-critical applications, it is of crucial importance that the predictive distribution of a classification model honestly expresses its true predictive uncertainty. In statistical terminology, such honest classifiers are called *reliable* or *calibrated* (see, e.g., Bröcker 2009; Zadrozny and Elkan 2002). Mathematically the reliability condition reads as

$$\mathbb{P}[Y \in \cdot \mid g(X)] = g(X), \tag{1}$$

---
[1]The code is in preparation to be outsourced for public use and will become available on https://github.com/uu-sml/calibration shortly.

i.e., the distribution of the target class conditional on any given prediction of our model is exactly equal to that prediction. For instance, the prediction $(0.95, 0.05)$ of a reliable binary classifier implies that the probability of the target being in the first class is 0.95 and being in the second class is 0.05, when considering all inputs resulting in the classifier output $(0.95, 0.05)$. Interestingly, within the machine learning community (Guo et al. 2017) a probabilistic model is called perfectly calibrated even if the much weaker condition

$$\mathbb{P}[Y = \arg\max g(X) \mid \max g(X)] = \max g(X) \quad (2)$$

is satisfied, possibly giving rise to some confusion. As discussed by Zadrozny and Elkan (2002), in practice it is also desired that

$$\mathbb{P}[Y = y \mid g(X)(\{y\})] = g(X)(\{y\}) \quad (3)$$

for all $y \in \mathcal{Y}$, i.e., that the marginal predictions of a model are calibrated. The conditions in Equations (1) to (3) are equivalent for binary classification problems but *not* in a general classification setting as the following example shows.

**Example 1 (A motivating toy example)**
Let us take a look at a simple example in which perfect calibration according to Guo et al. (2017) and calibrated marginal predictions do not imply reliability. Suppose $\mathcal{Y} = \{1, 2, 3\}$. Let $g$ be a probabilistic classifier that predicts the six probability distributions in the first column of Table 1 with equal probability and assume that the true conditional distribution $\mathbb{P}[Y \in \cdot \mid g(X)]$ is given by the second column. Figure 2 depicts these quantities on the probability simplex. The model is perfectly calibrated according to Guo et al. (2017) and additionally all marginal predictions are calibrated. However, $g$ is not reliable since (1) is not satisfied. Taking a safety-critical viewpoint and viewing model reliability as being of crucial importance, in this paper we set out to develop a comprehensive calibration evaluation methodology, a goal that will also require more future research to reach.

**Structure of the paper**  In Section 3, we present a general theoretical calibration evaluation framework allowing for investigation of arbitrary aspects of model calibration. Section 4 is dedicated to empirical evaluation of model calibration. It includes a study of estimators of key quantities, explanations of inherent subtleties of practical importance, refined interpretations of existing calibration evaluation techniques, and newly proposed calibration evaluation tools such as hypotheses tests and multidimensional reliability diagrams shown in Figure 1. The proposed evaluation methods are illustrated on standard neural network classifiers as well as on a tractable Gaussian mixture model in Section 5. The main text is supported by closely related appendices, available in the supplementary material, in which proofs, additional examples, explanations, and numerical results are included.

## 2.1 Related work

Practical application of calibrated classification models requires the knowledge of at least two crucial things: 1) how to evaluate the calibration of a given model, and 2) how to build a candidate model.

**Predictive uncertainty evaluation**  Calibration evaluation of neural networks in classification tasks has been a research topic for over a decade and has witnessed a resurgence of interest over the last two years. Niculescu-Mizil and Caruana (2005) used classical reliability diagrams (Murphy and Winkler 1977; Murphy and Winkler 1987) to investigate calibration of shallow neural networks in a binary classification setting and concluded that these networks are well-calibrated. Recently Guo et al. (2017) showed empirically that modern neural networks are ill-calibrated and ignited a strand of research (Kumar, Sarawagi, and Jain 2018; Zhang et al. 2018; Kendall and Gal 2017; Tran et al. 2018) proposing methods for obtaining calibrated classifiers. In all these works, calibration of the classifier is judged based on classical reliability diagrams and the so-called expected calibration error (ECE) (Guo et al. 2017) whose real-valued estimate is calculated from a reliability diagram and its associated confidence histogram. As we explain in this paper, calibration evaluation using this popular methodology suffers from certain shortcomings that we address in this work.

Another recent paper by Lakshminarayanan, Pritzel, and Blundell (2017) argues that predictive uncertainty calibration should be evaluated in terms of empirical approximations of expected scores based on strictly proper scoring rules (Gneiting and Raftery 2007). However, expected scores do not quantify model calibration directly; in fact, a model that is *not* close to the true classifier with respect to the expected score can nevertheless be calibrated. Also, different scoring rules can rank the same models differently (Winkler and Murphy 1968; Merkle and Steyvers 2013).

**Building calibrated probabilistic classifiers**  There exist two general approaches to obtain calibrated classifiers. The first one encourages model calibration already during training (Lakshminarayanan, Pritzel, and Blundell 2017; Kumar, Sarawagi, and Jain 2018; Zhang et al. 2018; Kendall and Gal 2017; Tran et al. 2018, see, e.g., ). An alternative approach is to recalibrate an existing ill-calibrated model; for binary classification this post-processing was applied by Platt (2000) and Zadrozny and Elkan (2001), and for multiclass classification by Zadrozny and Elkan (2002) and Guo et al. (2017). A quality comparison of different approaches calls for a sound calibration evaluation methodology. Moreover, reliability diagrams and expected miscalibration estimators can also be utilized as recalibration devices, similar to the work of Platt (2000) and Zadrozny and Elkan (2001), and hence the developments in the present paper are also of relevance for recalibration of classifiers.

| $g(X)$ | $\mathbb{P}[Y \in \cdot \mid g(X)]$ |
|---|---|
| (0.1, 0.3, 0.6) | (0.2, 0.2, 0.6) |
| (0.1, 0.6, 0.3) | (0.0, 0.7, 0.3) |
| (0.3, 0.1, 0.6) | (0.2, 0.2, 0.6) |
| (0.3, 0.6, 0.1) | (0.4, 0.5, 0.1) |
| (0.6, 0.1, 0.3) | (0.7, 0.0, 0.3) |
| (0.6, 0.3, 0.1) | (0.5, 0.4, 0.1) |

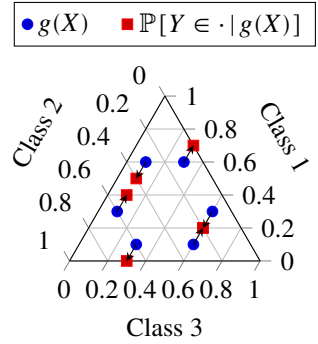

Table 1 & Figure 2: Probabilistic classifier $g$ for $\mathcal{Y} = \{1, 2, 3\}$ with six uniformly distributed predictions.

## 3 Theoretical framework

Before delving into calibration evaluation questions let us point out that for a given classification problem there are typically infinitely many calibrated models. Hence, there is hope to find a calibrated classifier (or at least one close to being calibrated) even though the ideal model $g^{\text{opt}}$ is almost impossible to find.

**Proposition 1 (Many calibrated classifiers).** *For any measurable function $h\colon \mathcal{X} \to \mathcal{Z}$, where $\mathcal{Z}$ is some measurable space, the map $g$ defined by*

$$g(X) := \mathbb{P}[Y \in \cdot \mid h(X)]$$

*is a calibrated probabilistic classifier.*

A proof of this statement is provided in Appendix A.2. In particular, if $h$ is a bijection, then $g$ is the ideal model $g^{\text{opt}}$. At the other extreme, if $h$ is a constant function, then $g$ is a calibrated but quite uninformative classifier that always predicts the marginal distribution of $Y$ regardless of the input.

### 3.1 Calibration functions

All information about the calibration of a probabilistic classifier $g$ is contained in its *(canonical) calibration function $r$* given by

$$r(\mu) := \mathbb{P}[Y \in \cdot \mid g(X) = \mu] \quad (4)$$

for every $\mu \in \mathcal{P}(\mathcal{Y})$ for which it is well-defined. By the definition of reliability in (1), a model is reliable if and only if its calibration function is the identity function. Furthermore, deviations from the identity function capture the degree of miscalibration present at different predictions. In many practical situations, especially when the number of target classes is large, we will typically not have enough data to accurately estimate the calibration function. Nonetheless, with the data available we might still be able to investigate certain aspects of model calibration; we next turn to formalize this idea of partial calibration evaluation.

The main principle underlying partial calibration evaluation is that a calibrated classifier induces calibrated classifiers for induced problems. Let $\psi\colon \mathcal{Y} \times \mathcal{P}(\mathcal{Y}) \to \mathcal{Z}$ be a measurable function that maps a pair of an observation in $\mathcal{Y}$ and a probabilistic prediction on $\mathcal{Y}$ to an element in a measurable space $\mathcal{Z}$ which is possibly different from $\mathcal{Y}$. Then $\psi$ induces a function $\pi_\psi\colon \mathcal{P}(\mathcal{Y}) \to \mathcal{P}(\mathcal{Z})$ given by

$$\pi_\psi(\mu) := \mathbb{P}_{Y' \sim \mu}[\psi(Y', \mu) \in \cdot], \quad \mu \in \mathcal{P}(\mathcal{Y}),$$

which maps a probabilistic prediction on $\mathcal{Y}$ to a prediction on $\mathcal{Z}$. Thus the model $g$ and the function $\psi$ yield an *induced predictive model* $g_\psi := \pi_\psi \circ g$ for an *induced problem*. In this induced problem, the random variable pairs $(X_1, Z_1), \ldots, (X_n, Z_n) \in \mathcal{X} \times \mathcal{Z}$, where $Z_i := \psi(Y_i, g(X_i))$, correspond to the observable data, and finding a probabilistic prediction of $Z = \psi(Y, g(X))$ given $X$ is of interest. Instead of the calibration function of $g$ in (4), we can consider the calibration function of $g_\psi$, which we also call the *calibration function $r_\psi\colon \mathcal{P}(\mathcal{Z}) \to \mathcal{P}(\mathcal{Z})$ induced by $\psi$*. It is given by

$$\begin{aligned}r_\psi(\nu) &:= \mathbb{P}[Z \in \cdot \mid g_\psi(X) = \nu] \\ &= \mathbb{P}[\psi(Y, g(X)) \in \cdot \mid \pi_\psi(g(X)) = \nu]\end{aligned}$$

for every $\nu \in \mathcal{P}(\mathcal{Z})$ for which it is well-defined. Informally, each function $\psi$ provides a lens through which we can inspect a particular aspect of model calibration. If a model is calibrated, then every induced calibration function is equal to the identity function. However, even if an induced calibration function is equal to the identity function, a model does not have to be calibrated, as can be seen from Example 1 together with Example 2 below. We now look at a few natural choices of the *calibration lens $\psi$*. Specifically, any popular calibration evaluation performed in machine learning literature can be phrased in this general setup.

**Example 2 (Calibration functions)**

(i) *The canonical calibration function.* The calibration function induced by $\psi\colon (y, \mu) \mapsto y$ is just the canonical calibration function.

(ii) *The fixed partition.* Fix a partition $\{A_i\}_{i=1}^k$ of $\mathcal{Y}$. Then $\psi\colon (y, \mu) \mapsto \sum_{i=1}^k i \mathbb{1}_{A_i}(y)$ induces a probabilistic classifier for a classification problem with target classes $A_1, \ldots, A_k$ and a calibration function that captures its reliability.

(iii) *The $k$ largest predictions.* Define $\alpha \colon \{1, \ldots, m\} \times \mathcal{P}(\mathcal{Y}) \to \mathcal{Y}$ such that $\alpha(i, \mu)$ is the class of the $i$th largest prediction according to $\mu$. Let $1 \leq k < m$ and define $\psi_k(y, \mu) := \sum_{i=1}^{k} i \mathbb{1}_{\{\alpha(i,\mu)\}}(y)$, which induces a calibration function capturing how well the $k$ largest predictions are calibrated. The perfect calibration definition by Guo et al. (2017) corresponds to the special case $k = 1$.

Since $\mathcal{Y}$ is a finite discrete space in classification problems, it is sufficient to consider only finite spaces $\mathcal{Z}$ with $|\mathcal{Z}| \leq m$. Consequently, every result about canonical calibration functions generalizes immediately to induced calibration functions by applying the results to the induced models. Thus without loss of generality we only provide results for canonical calibration functions.

### 3.2 Measures of miscalibration

Often it is desirable to summarize the overall calibration performance in a single number. This can be achieved with the help of a distance function $d \colon \Delta_{m-1} \times \Delta_{m-1} \to [0, \infty)$ gauging the closeness between the output of a probabilistic classifier and the corresponding value of the calibration function.

One natural quantity of interest is the *expected miscalibration on $A \subseteq \Delta_{m-1}$ with respect to $d$* defined by

$$\eta_d := \mathbb{E}[d(r(g(X)), g(X)) \mid g(X) \in A]. \quad (5)$$

Another interesting metric is the worst case miscalibration $\max_{\mu \in A} d(r(\mu), \mu)$. Choosing $A = \Delta_{m-1}$ allows us to quantify miscalibration on the entire probability simplex whereas by selecting $A \subsetneq \Delta_{m-1}$ we can focus on miscalibration in a particular region of interest. For the sake of brevity we will omit the dependency on the subset $A$ in our notation when no confusion arises.

In the language of our framework, Guo et al. (2017) used the popular total variation (TV) distance $d(x, y) = \frac{1}{2}\|x - y\|_1$ to measure miscalibration of an induced binary probabilistic classifier. The squared Euclidean distance $d(x, y) = \|x - y\|_2^2$ is another easily interpretable distance, though any distance function can be chosen to suit the application.

## 4 Empirical calibration evaluation

Having introduced the theoretical framework, we next turn to empirical evaluation questions.

### 4.1 Estimators and their properties

**Calibration functions** Let $\Phi = \{\Phi^{(i)}\}_{i=1}^{L}$ be a random partition of $A \subseteq \Delta_{m-1}$ with the randomness arising solely from the dependence on the observable data $\{(g(X_i), Y_i)\}_{i=1}^{n}$. We denote the unique set in this partition containing a vector $w \in A$ by $\Phi[w]$.

The histogram regression estimator $\hat{r} \colon A \to \mathcal{P}(\mathcal{Y}) \cup \{0\}$ of the calibration function on $A$ is defined by

$$\hat{r}(w)(\{y\}) := \frac{|\{i \colon g(X_i) \in \Phi[w] \wedge Y_i = y\}|}{|\{i \colon g(X_i) \in \Phi[w]\}|}$$

for all $w \in A$ and $y \in \mathcal{Y}$, with the convention that $\hat{r}(w) = 0$ if the denominator happens to be zero, i.e., if there are no data points in $\Phi[w]$. The estimate $\hat{r}$ is constant on every set in the partition $\Phi$, and hence we can define $\hat{r}^{(i)}$ as its unique value on $\Phi^{(i)}$.

For sensible partitioning schemes that result in finer and finer partitions as the data set grows the estimate $\hat{r}$ converges to the calibration function in an appropriate sense as shown by Nobel (1996). Splitting the probability simplex into bins of equal size (as often done in the machine learning literature for binary problems) is the simplest option. However, choosing a data-dependent binning scheme is known to provide much faster convergence in practice (Nobel 1996).

**Expected miscalibration** Let us denote the proportion and the average of all predictions in a subset $\Phi^{(i)}$ by

$$\hat{p}^{(i)} := \frac{|\{j \colon g(X_j) \in \Phi^{(i)}\}|}{|\{j \colon g(X_j) \in A\}|} \text{ and } \hat{g}^{(i)} := \frac{\sum_{j \colon g(X_j) \in \Phi^{(i)}} g(X_j)}{|\{j \colon g(X_j) \in \Phi^{(i)}\}|},$$

respectively. Similarly as before, we adopt the convention that $\hat{p}^{(i)} = 0$ and $\hat{g}^{(i)} = 0$ if the denominators happen to be zero.

The next result tells us that

$$\hat{\eta}_d := \sum_{i=1}^{L} \hat{p}^{(i)} d(\hat{r}^{(i)}, \hat{g}^{(i)}) \quad (6)$$

can be interpreted as an estimator of the expected miscalibration $\eta_d$. Guo et al. (2017) used a special case of this estimator to estimate the expected miscalibration of an induced binary classifier with respect to the total variation distance.

**Theorem 1.** *Suppose that the calibration function $r$ is Lipschitz continuous and $d \colon \Delta_{m-1} \times \Delta_{m-1} \to [0, \infty)$ is continuous and uniformly continuous in the first argument. Let $\{\Phi_N\}_{N \in \mathbb{N}}$ be a sequence of finite data-independent partitions of $A \subseteq \Delta_{m-1}$ such that $\lim_{N \to \infty} \max_{S \in \Phi_N} \operatorname{diam} S = 0$, where $\operatorname{diam} S := \sup_{x,y \in S} \|x - y\|_2$. Then*

$$\lim_{N \to \infty} \lim_{n \to \infty} \hat{\eta}_{d,N} = \eta_d,$$

*with limits in the almost sure sense, where estimator $\hat{\eta}_{d,N}$ is defined according to (6) for each $N \in \mathbb{N}$.*

Interestingly, the estimator in (6) yields a lower bound for the expected miscalibration if the partition is kept fixed as the size of the data set grows to infinity.

**Theorem 2.** *Let $d\colon \Delta_{m-1} \times \Delta_{m-1} \to [0, \infty)$ be a continuous convex function and let $\Phi$ be a finite data-independent partition of $A \subseteq \Delta_{m-1}$. Then*

$$\lim_{n\to\infty} \hat{\eta}_d \leq \eta_d, \quad (7)$$

*with limit in the almost sure sense. Moreover, if $d$ can be written as $d(p, p') = f(p - p')$ for a convex function $f$, then equality holds if and only if for every $\Phi^{(i)}$ there exists a set $S \subseteq \mathbb{R}^m$ such that $\mathbb{P}[r(g(X)) - g(X) \in S \mid g(X) \in \Phi^{(i)}] = 1$ and $f$ coincides almost surely with an affine function on the convex hull of $S$.*

Proofs of Theorems 1 and 2 are given in Appendix A.2. Note that the popular distances arising from $\|\cdot\|_1$ and $\|\cdot\|_2^2$ satisfy the convexity condition in Theorem 2.

### 4.2 Quantifying the estimator variability

In the previous section, we discussed estimation of the calibration function and the expected miscalibration. Abstractly, both can be viewed as a realization of a random variable of the form $h(D)$, where $D := \{(g(X_i), Y_i)\}_{i=1}^n$ is random and $h$ is a deterministic function. Thus to avoid being fooled by randomness, the distribution of the calculated statistic must be taken into account when drawing any conclusions.

**Consistency resampling** The randomness of the estimator $h(D)$ can be attributed to the randomness of $g(X_1), \ldots, g(X_n)$ and the randomness of $Y_1 \mid g(X_1), \ldots, Y_n \mid g(X_n)$. The variability of $h(D)$ due to the randomness of $g(X_i)$ can be estimated by bootstrapping new data sets $s^1, s^2, \ldots$, where $s^j = \{g_i^j\}_{i=1}^n$, from the predictions $\{g(x_i)\}_{i=1}^n$ (see Efron and Hastie 2016). Moreover, assuming that the true calibration function $r$ equals some known function $\rho$ (in the following we indicate this assumption with a superscript $\rho$), we can approximate the variability of the estimator $h(D^\rho)$ due to the randomness arising from $Y_i^\rho \mid g(X_i)$ by sampling values $y_1^{\rho,j}, \ldots, y_n^{\rho,j}$ from the distributions $\rho(g_1^j), \ldots, \rho(g_n^j)$, respectively. The described resampling procedure is referred to as *consistency resampling* (Bröcker and Smith 2007). Given a big enough data set, we can also perform the usual bootstrapping from the complete data set $\underline{D} := \{(g(x_i), y_i)\}_{i=1}^n$ to estimate the variability of $h(D)$.

### 4.3 Interpreting empirical calibration evaluation

**Testing a hypothesis of perfect calibration** The approximate distribution of $h(D^\rho)$ obtained via consistency resampling can be used to perform null hypothesis tests. A natural choice is to check whether the calibration function equals the identity function, i.e., $\rho = \text{id}$, as is the case for calibrated models. If $h$ is a real-valued function we can estimate $\mathbb{P}[h(D^{\text{id}}) \geq h(\underline{D})]$ and use this value as a p-value to reject the hypothesis of perfect calibration.

**Comparing miscalibration estimates** Suppose we want to compare two probabilistic classifiers $g$ and $g'$ in terms of their reliability. Let $\hat{\eta}_d$ and $\hat{\eta}'_d$ be the estimators of $\eta_d$ and $\eta'_d$, respectively. It is currently common practice (see, e.g., Guo et al. 2017; Kumar, Sarawagi, and Jain 2018; Naeini, Cooper, and Hauskrecht 2015) to compare the reliability of $g$ and $g'$ in terms of just the realized values $\hat{\underline{\eta}}_d$ and $\hat{\underline{\eta}}'_d$ of $\hat{\eta}_d$ and $\hat{\eta}'_d$, respectively. Alas, we do not know anything about how much the biases $\mathbb{E}[\hat{\eta}_d] - \eta_d$ and $\mathbb{E}[\hat{\eta}'_d] - \eta'_d$ differ (it is easy to come up with examples showing that the two biases can differ significantly, see Appendix A.1). Moreover, the statistics $\hat{\eta}_d$ and $\hat{\eta}'_d$ are random variables typically having different distributions. As a result, comparing calibration of two models just in terms of the realizations of these estimators is unjustified and not necessarily meaningful. This reasoning indicates that the current widespread practice does not reveal the full story and leads to certain doubts about the conclusions reached.

Let us see how the direct comparison in terms of the estimates contrasts with our preferred p-value approach. Using the ideas of Section 4.2, we can approximate the distributions of $\hat{\eta}_d^{\text{id}}$ and $\hat{\eta}_d^{\prime\text{id}}$ in order to calculate $\mathbb{P}[\hat{\eta}_d^{\text{id}} \geq \hat{\underline{\eta}}_d]$ and $\mathbb{P}[\hat{\eta}_d^{\prime\text{id}} \geq \hat{\underline{\eta}}'_d]$. It might be that $\hat{\underline{\eta}}_d > \hat{\underline{\eta}}'_d$ but $\mathbb{P}[\hat{\eta}_d^{\text{id}} \geq \hat{\underline{\eta}}_d] > \alpha > \mathbb{P}[\hat{\eta}_d^{\prime\text{id}} \geq \hat{\underline{\eta}}'_d]$, where $\alpha$ is a chosen significance level. Hence, our p-value test would reject the hypothesis of perfect calibration for model $g'$ based on the small estimate $\mathbb{P}[\hat{\eta}_d^{\prime\text{id}} \geq \hat{\underline{\eta}}'_d] < \alpha$ but not for model $g$, regardless of the larger estimate $\hat{\underline{\eta}}_d > \hat{\underline{\eta}}'_d$.

**Lower bound interpretation** The result in (7) tells us that $\hat{\eta}_d$ calculated using a fixed partition is prone to underestimating the true expected miscalibration for uncalibrated models as the size of the data set grows to infinity. The condition for strict inequality in Theorem 2 holds for distances arising from strictly convex functions (unless $r(g(X)) - g(X)$ is constant within every bin) as well as for the total variation distance used by Guo et al. (2017) (unless $r(g(X)) - g(X)$ belongs to a single orthant within every bin). Hence, the common practice of estimating expected miscalibration with a fixed small number of bins and many data points can lead to reporting better than actual model miscalibration measures and thus might even put the safety of machine learning applications at risk. Rephrased in statistical terminology, the popular estimator in (6) for the expected miscalibration is asymptotically inconsistent in many situations with a negative asymptotic bias.

We emphasize, however, that this lower bound interpretation relies on asymptotics in the number of data points with a fixed binning. Establishing rates of convergence for the bias and variance of various binning schemes is an interesting topic for future work, which would enable us to draw more informative conclusions about the risk of underestimating miscalibration.

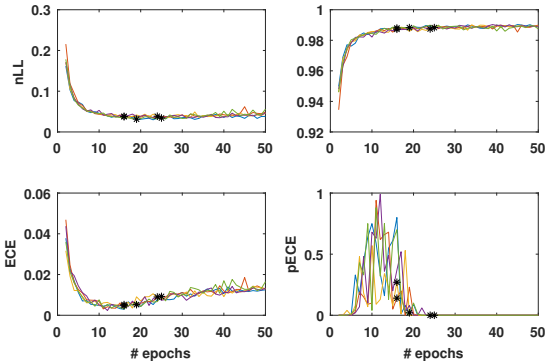

Figure 3: Negative log-likelihood (nLL), accuracy, expected miscalibration (ECE), and p-value under consistency assumption (pECE) of LeNet during training on MNIST (5 different initializations). Asteriks indicate models with minimum negative log-likelihood on the validation data set.

### 4.4 Visualizing model miscalibration

**Classical reliability diagrams** A classical reliability diagram is a method to investigate the miscalibration of a binary probabilistic classifier (see Murphy and Winkler 1977; Bröcker 2008, for an early and a more recent account) by plotting for one class the average prediction versus the regression estimate of the calibration function in each bin. In a binary classification setting the reliability diagram contains all information about $\hat{g}^{(i)}$ and $\hat{r}^{(i)}$ for every $i$ and hence can be viewed as depicting estimates for $\mathbb{E}[g(X) \,|\, g(X) \in \Phi^{(i)}]$ and $\mathbb{E}[r(g(X)) \,|\, g(X) \in \Phi^{(i)}]$.

In Figure 4, we present our variant of the classical reliability diagram inspired by Bröcker and Smith (2007). We plot the difference between the regression estimate and the average prediction (the so-called deviation) instead of just the regression estimate. In our opinion, this modification simplifies the visual comparison of both estimates while retaining the same information. Additionally consistency bars indicate quantiles of the estimates obtained under the consistency assumption, either by resampling or by making use of the binomial distribution of consistent outcomes as discussed by Bröcker and Smith (2007). Thus consistency bars provide a visual indication of the range within which deviations are likely to appear even if the model is calibrated. Moreover, they can be interpreted as a hypothesis test of perfect calibration in each bin. Figure 4a shows equally-sized bins whereas Figure 4b presents the same data using a data-dependent binning scheme that is explained in Appendix B.

**Higher-dimensional reliability diagrams** In multiclass classification, a classical one-dimensional reliability diagram contains information only about a specific partial aspect of model reliability as it only represents an induced binary classification problem (recall Example 1 and Section 3.1). A direct generalization of our one-dimensional reliability diagram to higher dimensions would be to plot the deviations of the regression estimates from the average predictions for all but one class as arrows pointing from the average predictions to the regression estimates. However, such a visualization would depend on the choice of the left-out class. In our opinion, a better alternative is to plot the deviations of the estimates from the average predictions of all classes as arrows directly on the probability simplex. An estimate of the proportion of predictions falling into a bin can be included in the plot by color-coding the corresponding arrows or bins.

An example of a higher-dimensional reliability diagram is presented in Figure 1. More details such as the variability of the estimates (similar to consistency bars in one-dimensional reliability diagrams) can be added to the diagrams but are not included here to avoid overloaded visualizations. For classification problems with up to four classes, a full multiclass classification reliability diagram can be plotted. If the number of classes is greater than four, we cannot visualize the full probability simplex, but we can still explore particular aspects of model calibration using one-, two-, and three-dimensional reliability diagrams of induced classifiers.

## 5 Experimental results

Next, we illustrate the proposed calibration evaluation framework with reliability diagrams by applying it to an analytically tractable example and standard neural networks.

### 5.1 One-dimensional Gaussian mixture model

We consider a simple Gaussian mixture model consisting of two equally likely standard normally distributed classes with mean $-1$ and $1$. More concretely, we take $\mathcal{X} = \mathbb{R}$, $\mathcal{Y} = \{-1, 1\}$, $p(Y = -1) = p(Y = 1) = 1/2$, and $p(X = x \,|\, Y = y) = \mathcal{N}(x; y, 1)$ for all $(x, y) \in \mathcal{X} \times \mathcal{Y}$. This yields $p(Y = -1 \,|\, X = x) = 1/(1 + \exp(2x))$, and thus the perfect model is covered by the class of logistic regression models $g \colon \mathcal{X} \to \Delta_1$ of the form $g_1(x) = 1/(1+\exp(-(\beta_0 + \beta_1 x)))$, $g_2(x) = 1 - g_1(x)$, with parameters $\beta_0, \beta_1 \in \mathbb{R}$. The only calibrated logistic regression models are the perfect model ($\beta_0 = 0$, $\beta_1 = -2$) and the constant model with parameters $\beta_0 = \beta_1 = 0$ and $g_1(x) = g_2(x) = 1/2$ for all $x$. In addition to those calibrated models, we consider the uncalibrated logistic regression model with $\beta_0 = \beta_1 = 1$.

The expected miscalibration $\eta_{\text{TV}}$ with respect to the total variation distance is 0 for the calibrated models and approximately 0.56 for the uncalibrated model. For each of these models we repeatedly computed empirical estimates of miscalibration according to (6) for varying number of samples and bins. The results are presented in Figures 6, 7, 9, 10, 12 and 13 in Appendix C. As expected, for the calibrated models the empirical estimates are close to zero but almost always overestimate the expected miscalibration. For the uncalibrated model both over- and underestimation of the expected miscalibration can be observed. It seems that in this example with increasing number of samples and, maybe even more importantly, increasing number of samples per bin, underestimation becomes more likely.

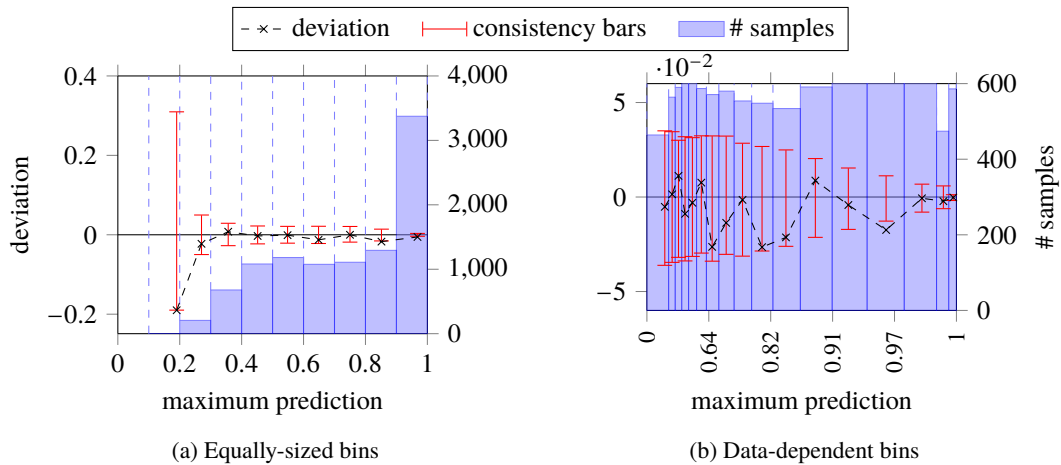

(a) Equally-sized bins

(b) Data-dependent bins

Figure 4: Reliability diagrams for the maximum predictions of LeNet on the CIFAR-10 test set w.r.t. the total variation distance. Crosses indicate the deviation of the outcome distribution from the predictions in each bin. Blue bars show the distribution of predictions. Red bars visualize the 5th and 95th percentiles of the deviation in 1000 consistency resamples.

Moreover, we randomly generated 10000 input-output pairs of the Gaussian mixture model and plotted one-dimensional reliability diagrams for the perfect model, the calibrated constant model, and the uncalibrated model in Figures 8, 11 and 14 in Appendix C, respectively. In addition to the empirical deviation, the distribution of the predictions, and the consistency bars, these plots show the true analytical deviation, which is closely matched by the empirical estimates.

### 5.2 Neural networks

We trained LeNet (LeCun et al. 1998), DenseNet (Huang et al. 2017), and ResNet (He et al. 2016) models on the CIFAR-10 data set (Krizhevsky 2009) in the standard way described in the literature. For the LeNet model, one-dimensional reliability diagrams of the maximum predictions are shown in Figure 4, for equally-sized and data-dependent bins. Visually both diagrams cannot rule out the reliability hypothesis for the LeNet model, in accordance with previous publications (Guo et al. 2017; Niculescu-Mizil and Caruana 2005).

Various alternative inspections of reliability are possible with the proposed two-dimensional reliability diagrams such as the plots in Figure 1, for which the original 10 classes of the CIFAR-10 data set are combined into three groups. In these diagrams, the deviation between outcomes and average predictions is small, particularly in regions with frequent predictions, which is again consistent with a calibrated model hypothesis. The one- and two-dimensional reliability diagrams of the DenseNet and ResNet models in Figures 15 to 18 in Appendix C, and in particular the one-dimensional reliability diagrams with data-dependent bins, do not seem to support the reliability hypothesis for these models to the same extent, in line with previous results by Guo et al. (2017).

To observe the proposed p-value test in action, we trained LeNet on the MNIST data set (LeCun et al. 1998) and visualized the p-value evolution as well as other relevant metrics during training in Figure 3. Interestingly, we see that models with the best predictive uncertainty as measured by the negative log-likelihood (a strictly proper scoring rule evaluation advocated by Lakshminarayanan, Pritzel, and Blundell (2017)) are quite consistently indicated as miscalibrated by the p-value test. Moreover, in Appendix D, expected miscalibration estimates are calculated for different neural networks. Our results exhibit the importance of the binning scheme and show that ResNet and DenseNet can have lower expected miscalibration estimates than LeNet, contrasting with the model reliability story told by the reliability diagrams.

## 6 Conclusion

Evaluation of model calibration is about checking whether probabilities predicted by a model match the distribution of realized outcomes. In this article, we built on existing calibration evaluation approaches and proposed a general mathematical framework for evaluating model calibration, or a chosen aspect of it, in classification problems. We showed that empirical estimates of intuitive miscalibration measures should not be used in a naive way to compare probabilistic classifiers but instead can be employed in hypothesis tests for testing model reliability. We hope our developments and attempts in rigorous model calibration evaluation will encourage other researchers to study this essential topic further.

# Supplementary material for 'Evaluating model calibration in classification'

## A  Theoretical results on calibration evaluation

### A.1  Additional examples

The following example shows that perfect calibration according to Guo et al. (2017) does not imply calibrated marginal predictions.

**Example 3 (No calibrated marginal predictions)**
Suppose $\mathcal{Y} = \{1, 2, 3\}$. Let $g$ be a probabilistic classifier that predicts only the two probability distributions in the first column of Table 2 with equal probability and assume that the true conditional distribution $\mathbb{P}[Y \in \cdot \,|\, g(X)]$ is given by the second column. The model is perfectly calibrated according to Guo et al. (2017). However, all marginal predictions are uncalibrated and additionally $g$ is not reliable since (1) is not satisfied. Figure 5 provides an illustration of these observations.

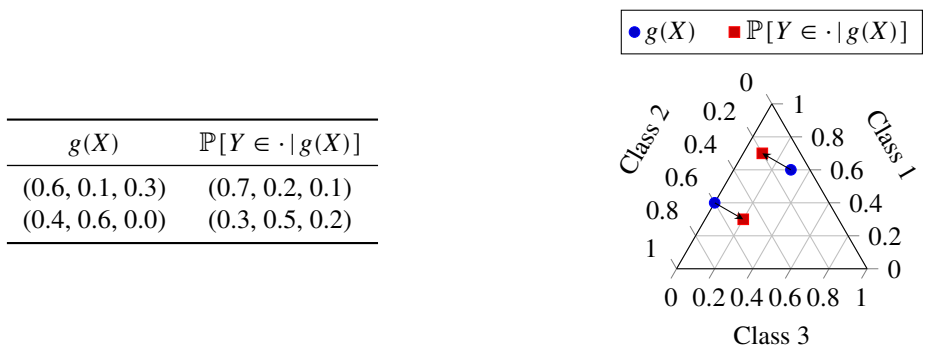

| $g(X)$ | $\mathbb{P}[Y \in \cdot \,|\, g(X)]$ |
|---|---|
| (0.6, 0.1, 0.3) | (0.7, 0.2, 0.1) |
| (0.4, 0.6, 0.0) | (0.3, 0.5, 0.2) |

Table 2 & Figure 5: Probabilistic classifier $g$ for $\mathcal{Y} = \{1, 2, 3\}$ with two uniformly distributed predictions.

Using a modification of our framework, one can express the calibration evaluation by Kendall and Gal (2017) by an induced calibration function.

**Example 4 (Marginalized calibration)**
Kendall and Gal (2017) evaluated model calibration by estimating the calibration function

$$r_{\psi}(u) = \mathbb{P}[\psi_u(Y, \mu) \in \cdot \,|\, \pi_{\psi_u}(g(X)) = u\delta_1 + (1 - u)\delta_0]$$

induced by $\psi_u(y, \mu) = \mathbb{1}_{\{\mu(\{y\}) = u\}}$, where $u \in [0, 1]$ and $\delta_a$ denotes the Dirac measure at $a$.

The following example shows that the biases of the estimators of expected miscalibration can differ significantly even for calibrated models.

**Example 5 (Different bias)**
Consider a binary classification problem with $\mathcal{Y} = \{1, 2\}$ and $p(Y = 1) = p(Y = 2) = 1/2$ that is clearly separable, i.e., with $p(Y = 0 \,|\, X = x) \in \{0, 1\}$ for all $x \in \mathcal{X}$. Both the optimal model $g^{\text{opt}}$ and the constant model $g \equiv (1/2, 1/2)$ are calibrated and hence the expected miscalibration $\eta_{\text{TV}}$ is zero for both models.

Let us define an estimator $\hat{\eta}_{\text{TV}}$ of expected miscalibration with only one bin on the whole probability simplex and a single data point $(X, Y)$. Then the bias of the estimator is $\mathbb{E}[\hat{\eta}_{\text{TV}}] - \eta_{\text{TV}} = 0$ for the perfect model since the expected miscalibration estimate is always zero. However, the constant model yields a bias of $\mathbb{E}[\hat{\eta}_{\text{TV}}] - \eta_{\text{TV}} = 1/2$ since in that case the expected miscalibration estimate is always $1/2$.

### A.2  Proofs

**Proposition 1 (Many calibrated classifiers).** *For any measurable function $h\colon \mathcal{X} \to \mathcal{Z}$, where $\mathcal{Z}$ is some measurable space, the map $g$ defined by*

$$g(X) := \mathbb{P}[Y \in \cdot \,|\, h(X)]$$

*is a calibrated probabilistic classifier.*

*Proof.* Let $y \in \mathcal{Y}$. We have

$$g(X)(\{y\}) = \mathbb{E}[g(X)(\{y\}) \,|\, g(X)] = \mathbb{E}[\mathbb{E}[\mathbb{1}_{\{y\}}(Y) \,|\, h(X)] \,|\, g(X)].$$

Since by definition $g(X)$ is a function of $h(X)$ it follows from the tower property that

$$g(X)(\{y\}) = \mathbb{E}[\mathbb{1}_{\{y\}}(Y) \,|\, g(X)] = \mathbb{P}[Y = y \,|\, g(X)].$$

Hence $g(X) = \mathbb{P}[Y \in \cdot \,|\, g(X)]$, and therefore the probabilistic classifier $g$ is calibrated. $\square$

**Theorem 1.** *Suppose that the calibration function $r$ is Lipschitz continuous and $d\colon \Delta_{m-1} \times \Delta_{m-1} \to [0, \infty)$ is continuous and uniformly continuous in the first argument. Let $\{\Phi_N\}_{N \in \mathbb{N}}$ be a sequence of finite data-independent partitions of $A \subseteq \Delta_{m-1}$ such that $\lim_{N \to \infty} \max_{S \in \Phi_N} \operatorname{diam} S = 0$, where $\operatorname{diam} S := \sup_{x,y \in S} \|x - y\|_2$. Then*

$$\lim_{N \to \infty} \lim_{n \to \infty} \hat{\eta}_{d,N} = \eta_d,$$

*with limits in the almost sure sense, where estimator $\hat{\eta}_{d,N}$ is defined according to (6) for each $N \in \mathbb{N}$.*

*Proof.* To keep the notation simple we provide a proof for $A = \Delta_{m-1}$. The case $A \subsetneq \Delta_{m-1}$ follows in the same way by conditioning on $g(X) \in A$.

For $N \in \mathbb{N}$ let $\Phi_N = \{\Phi_N^{(i)}\}_{i=1}^{l_N}$ be a finite data-independent partition of $\Delta_{m-1}$ such that $\lim_{N \to \infty} \max_i \operatorname{diam} \Phi_N^{(i)} = 0$. We define $\hat{r}_N^{(i)}, \hat{g}_N^{(i)}, \hat{p}_N^{(i)}$, and $\hat{\eta}_{d,N}$ analogously to the notation in Section 4.1. Similarly we denote the average output distribution, the average predicted distribution, and the proportion of predictions in subset $\Phi_N^{(i)}$ by $\bar{r}_N^{(i)} := \mathbb{E}[r(g(X)) \,|\, g(X) \in \Phi_N^{(i)}]$, $\bar{g}_N^{(i)} := \mathbb{E}[g(X) \,|\, g(X) \in \Phi_N^{(i)}]$, and $\bar{p}_N^{(i)} := \mathbb{P}[g(X) \in \Phi_N^{(i)}]$, respectively.

From the continuous mapping theorem it follows that for all $N \in \mathbb{N}$

$$\lim_{n \to \infty} \sum_{i=1}^{l_N} \left| \hat{p}_N^{(i)} d\left(\hat{r}_N^{(i)}, \hat{g}_N^{(i)}\right) - \bar{p}_N^{(i)} d\left(r(\bar{g}_N^{(i)}), \bar{g}_N^{(i)}\right) \right| = \sum_{i=1}^{l_N} \bar{p}_N^{(i)} \left| d\left(\bar{r}_N^{(i)}, \bar{g}_N^{(i)}\right) - d\left(r(\bar{g}_N^{(i)}), \bar{g}_N^{(i)}\right) \right|, \tag{8}$$

with limit in the almost sure sense.

Let $K \geq 0$ be a Lipschitz constant for calibration function $r$. Hence for all $N \in \mathbb{N}$ and $i \in \{1, \ldots, l_N\}$ we have

$$\begin{aligned}
\left\| \bar{r}_N^{(i)} - r(\bar{g}_N^{(i)}) \right\|_2 &= \left\| \mathbb{E}\left[ r(g(X)) - r(\bar{g}_N^{(i)}) \,|\, g(X) \in \Phi_N^{(i)} \right] \right\|_2 \leq \mathbb{E}\left[ \|r(g(X)) - r(\bar{g}_N^{(i)})\|_2 \,|\, g(X) \in \Phi_N^{(i)} \right] \\
&\leq K \mathbb{E}\left[ \|g(X) - \bar{g}_N^{(i)}\|_2 \,|\, g(X) \in \Phi_N^{(i)} \right] \leq K \mathbb{E}\left[ \operatorname{diam} \Phi_N^{(i)} \,|\, g(X) \in \Phi_N^{(i)} \right] = K \operatorname{diam} \Phi_N^{(i)} \\
&\leq K \max_{S \in \Phi_N} \operatorname{diam} S.
\end{aligned} \tag{9}$$

Let $\epsilon > 0$. Since by assumption distance function $d$ is uniformly continuous in the first argument, there exists $\delta > 0$ such that for all $x, y, z \in \Delta_{m-1}$ with $\|x - y\|_2 < \delta$ inequality $|d(x, z) - d(y, z)| < \epsilon$ holds. From (9) and the assumption $\lim_{N \to \infty} \max_{S \in \Phi_N} \operatorname{diam} S = 0$ we know that there exists $N_0 \in \mathbb{N}$ such that for all $N \geq N_0$ and all $i \in \{1, \ldots, l_N\}$ we have $\|\bar{r}_N^{(i)} - r(\bar{g}_N^{(i)})\|_2 < \delta$. Hence together with (8) we obtain for all $N \geq N_0$

$$\lim_{n \to \infty} \sum_{i=1}^{l_N} \left| \hat{p}_N^{(i)} d\left(\hat{r}_N^{(i)}, \hat{g}_N^{(i)}\right) - \bar{p}_N^{(i)} d\left(r(\bar{g}_N^{(i)}), \bar{g}_N^{(i)}\right) \right| < \sum_{i=1}^{l_N} \bar{p}_N^{(i)} \epsilon = \epsilon,$$

with limit in the almost sure sense. Since $\epsilon$ was chosen arbitrarily this implies

$$\lim_{N \to \infty} \lim_{n \to \infty} \sum_{i=1}^{l_N} \left| \hat{p}_N^{(i)} d\left(\hat{r}_N^{(i)}, \hat{g}_N^{(i)}\right) - \bar{p}_N^{(i)} d\left(r(\bar{g}_N^{(i)}), \bar{g}_N^{(i)}\right) \right| = 0, \tag{10}$$

with limits in the almost sure sense.

For all $N \in \mathbb{N}$ the triangle inequality yields, with limits taken in the almost sure sense,

$$\begin{aligned}
\left| \eta_d - \lim_{n \to \infty} \hat{\eta}_{d,N} \right| &= \left| \mathbb{E}\left[ d(r(g(X)), g(X)) \right] - \lim_{n \to \infty} \sum_{i=1}^{l_N} \hat{p}_N^{(i)} d\left(\hat{r}_N^{(i)}, \hat{g}_N^{(i)}\right) \right| \\
&\leq \left| \mathbb{E}\left[ d(r(g(X)), g(X)) \right] - \sum_{i=1}^{l_N} \bar{p}_N^{(i)} d\left(r(\bar{g}_N^{(i)}), \bar{g}_N^{(i)}\right) \right| \\
&\quad + \left| \lim_{n \to \infty} \sum_{i=1}^{l_N} \hat{p}_N^{(i)} d\left(\hat{r}_N^{(i)}, \hat{g}_N^{(i)}\right) - \sum_{i=1}^{l_N} \bar{p}_N^{(i)} d\left(r(\bar{g}_N^{(i)}), \bar{g}_N^{(i)}\right) \right|.
\end{aligned} \tag{11}$$

From the definition of the Riemann-Stieltjes integral it follows that

$$\lim_{N \to \infty} \left| \mathbb{E}\left[d\left(r(g(X)), g(X)\right)\right] - \sum_{i=1}^{l_N} \overline{p}_N^{(i)} d\left(r(\overline{g}_N^{(i)}), \overline{g}_N^{(i)}\right) \right| = 0, \qquad (12)$$

and (10) implies that

$$\lim_{N \to \infty} \left| \lim_{n \to \infty} \sum_{i=1}^{l_N} \hat{p}_N^{(i)} d\left(\hat{r}_N^{(i)}, \hat{g}_N^{(i)}\right) - \sum_{i=1}^{l_N} \overline{p}_N^{(i)} d\left(r(\overline{g}_N^{(i)}), \overline{g}_N^{(i)}\right) \right|$$

$$\leq \lim_{N \to \infty} \lim_{n \to \infty} \sum_{i=1}^{l_N} \left| \hat{p}_N^{(i)} d\left(\hat{r}_N^{(i)}, \hat{g}_N^{(i)}\right) - \overline{p}_N^{(i)} d\left(r(\overline{g}_N^{(i)}), \overline{g}_N^{(i)}\right) \right| = 0, \qquad (13)$$

with limits in the almost sure sense. Thus all in all, from Equations (11) to (13) we obtain

$$\left| \eta_d - \lim_{N \to \infty} \lim_{n \to \infty} \hat{\eta}_{d,N} \right| = \lim_{N \to \infty} \left| \eta_d - \lim_{n \to \infty} \hat{\eta}_{d,N} \right| \leq 0 + 0 = 0,$$

with limits in the almost sure sense. $\square$

**Theorem 2.** *Let $d \colon \Delta_{m-1} \times \Delta_{m-1} \to [0, \infty)$ be a continuous convex function and let $\Phi$ be a finite data-independent partition of $A \subseteq \Delta_{m-1}$. Then*

$$\lim_{n \to \infty} \hat{\eta}_d \leq \eta_d, \qquad (7)$$

*with limit in the almost sure sense. Moreover, if $d$ can be written as $d(p, p') = f(p - p')$ for a convex function $f$, then equality holds if and only if for every $\Phi^{(i)}$ there exists a set $S \subseteq \mathbb{R}^m$ such that $\mathbb{P}[r(g(X)) - g(X) \in S \mid g(X) \in \Phi^{(i)}] = 1$ and $f$ coincides almost surely with an affine function on the convex hull of $S$.*

*Proof.* To keep the notation simple we provide a proof for $A = \Delta_{m-1}$. The case $A \subsetneq \Delta_{m-1}$ follows in the same way by conditioning on $g(X) \in A$.

Let $\Phi = \{\Phi^{(i)}\}_{i=1}^{l}$ be a finite data-independent partition of $\Delta_{m-1}$. Then we have

$$\eta_d = \mathbb{E}[d(r(g(X)), g(X))] = \sum_{i=1}^{l} \mathbb{P}[g(X) \in \Phi^{(i)}] \mathbb{E}[d(r(g(X)), g(X)) \mid g(X) \in \Phi^{(i)}] \qquad (14)$$

$$\geq \sum_{i=1}^{l} \mathbb{P}[g(X) \in \Phi^{(i)}] d\left(\mathbb{E}[r(g(X)) \mid g(X) \in \Phi^{(i)}], \mathbb{E}[g(X) \mid g(X) \in \Phi^{(i)}]\right), \qquad (15)$$

where (14) follows from the law of total probability and (15) from Jensen's inequality. Hence by the continuous mapping theorem we get

$$\eta_d \geq \lim_{n \to \infty} \sum_{i=1}^{l} \hat{p}^{(i)} d(\hat{r}^{(i)}, \hat{g}^{(i)}) = \lim_{n \to \infty} \hat{\eta}_d,$$

with limit in the almost sure sense. The exact equality condition is obtained by unwrapping the equality condition in Jensen's inequality. $\square$

## B  Binning schemes

With enough data available the calibration function can be approximated by partitioning the probability simplex into bins and calculating the observed empirical frequencies of realized outcomes associated with the bins, e.g., by using the histogram regression of Nobel (1996). In the commonly considered binary classification setting the unit interval [0, 1] is typically split into a given number of intervals of equal width (Bröcker 2008) (fixed-width binning). This approach can be extended to multiple dimensions by using symmetric equally-sized higher-dimensional bins but the number of bins grows exponentially with the number of classes. Additionally, predictions of neural networks after training are usually highly non-uniformly distributed, often making accurate estimation of the calibration function in multiclass classification with moderately sized amounts of data practically infeasible in large parts of the probability simplex.

Thus an attractive alternative is to partition the probability simplex into bins with approximately equal number of predictions. Bröcker (2008) suggests this binning procedure even in the case of non-uniformly distributed predictions of binary outcomes. In our study we employed a simple recursive partitioning scheme and split predictions along the mean of the dimension with highest variance as long as the number of predictions per bin was above a given threshold value, which was typically set to 1000 in our experiments. As discussed by Nobel (1996), different data-dependent binning schemes are possible and described in literature.

## C  Additional visualizations

### C.1  Gaussian mixture model

#### C.1.1  Perfect model

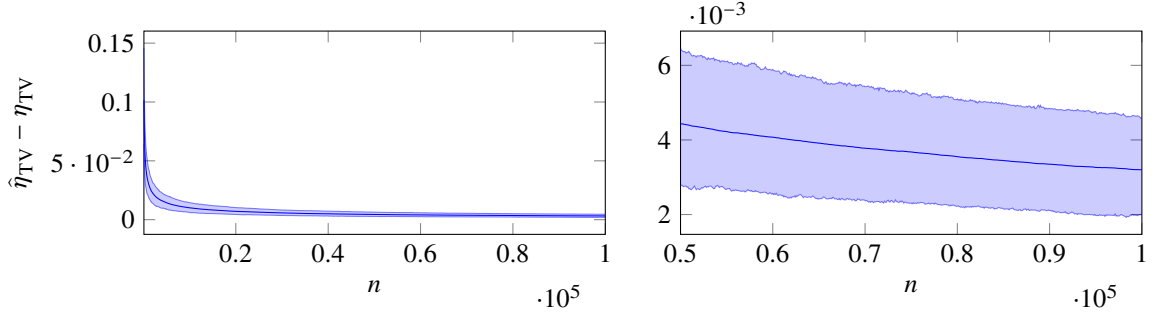

Figure 6: 5th percentile, mean, and 95th percentile of the difference of estimated and expected miscalibration of the perfect model ($\beta_0 = 0$, $\beta_1 = -2$) w.r.t. the total variation distance and 10 equally-sized bins (1000 series of random data).

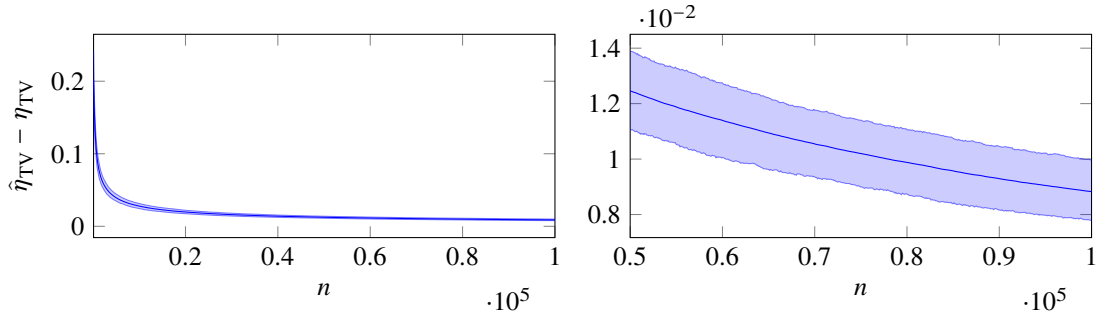

Figure 7: 5th percentile, mean, and 95th percentile of the difference of estimated and expected miscalibration of the perfect model ($\beta_0 = 0$, $\beta_1 = -2$) w.r.t. the total variation distance and 100 equally-sized bins (1000 series of random data).

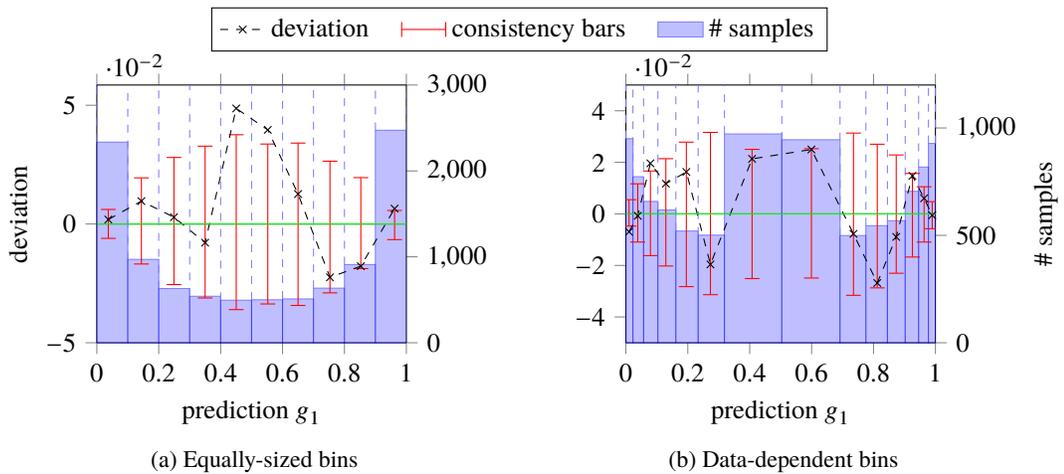

(a) Equally-sized bins

(b) Data-dependent bins

Figure 8: Reliability diagrams for the perfect model ($\beta_0 = 0$, $\beta_1 = -2$) w.r.t. the total variation distance on a randomly generated test set (10000 inputs). Crosses indicate the deviation of the outcome distribution from the predictions in each bin. Blue bars show the distribution of predictions. Red bars visualize the 5th and 95th percentiles of the deviation in 1000 consistency resamples. The green curve shows the true analytical deviation.

### C.1.2 Calibrated constant model

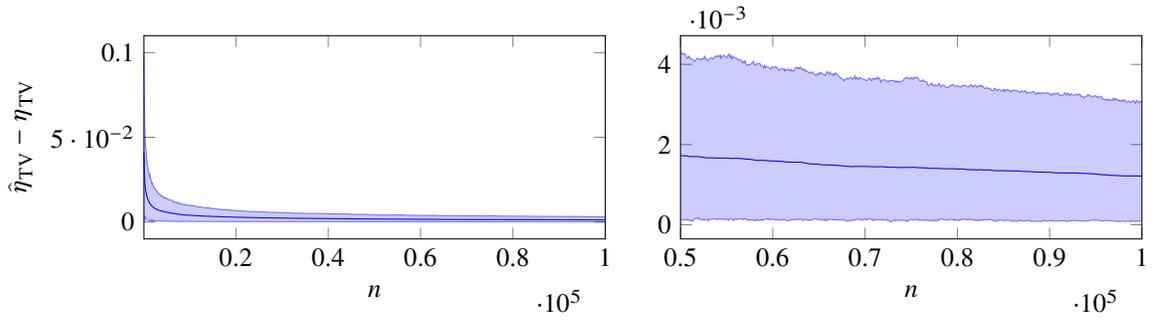

Figure 9: 5th percentile, mean, and 95th percentile of the difference of estimated and expected miscalibration of the calibrated constant model ($\beta_0 = \beta_1 = 0$) w.r.t. the total variation distance and 10 equally-sized bins (1000 series of random data).

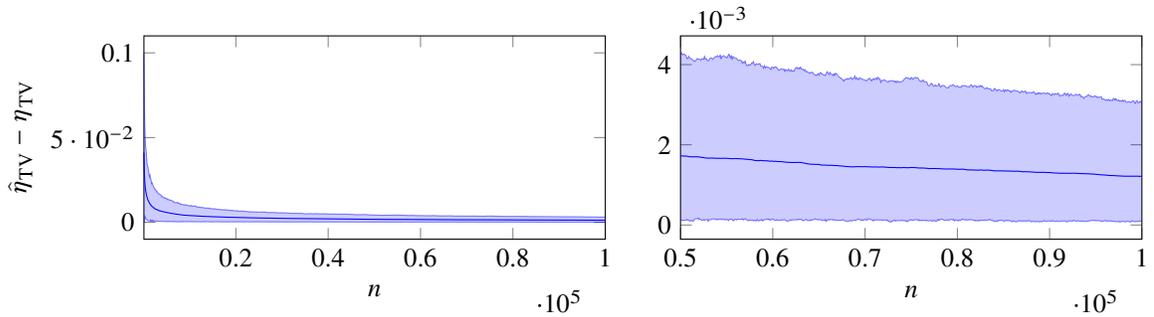

Figure 10: 5th percentile, mean, and 95th percentile of the difference of estimated and expected miscalibration of the calibrated constant model ($\beta_0 = \beta_1 = 0$) w.r.t. the total variation distance and 100 equally-sized bins (1000 series of random data).

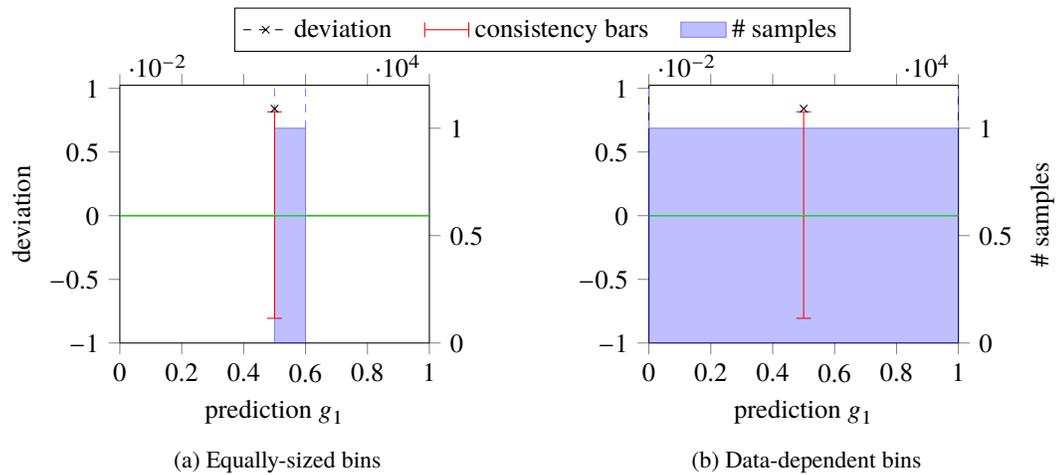

(a) Equally-sized bins

(b) Data-dependent bins

Figure 11: Reliability diagrams for the calibrated constant model ($\beta_0 = \beta_1 = 0$) w.r.t. the total variation distance on a randomly generated test set (10000 inputs). Crosses indicate the deviation of the outcome distribution from the predictions in each bin. Blue bars show the distribution of predictions. Red bars visualize the 5th and 95th percentiles of the deviation in 1000 consistency resamples. The green curve shows the true analytical deviation.

### C.1.3 Uncalibrated model

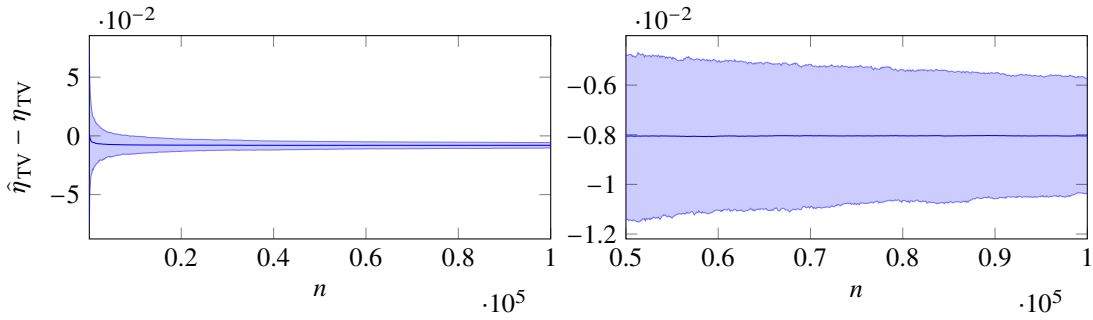

Figure 12: 5th percentile, mean, and 95th percentile of the difference of estimated and expected miscalibration of the uncalibrated model ($\beta_0 = \beta_1 = 1$) w.r.t. the total variation distance and 10 equally-sized bins (1000 series of random data).

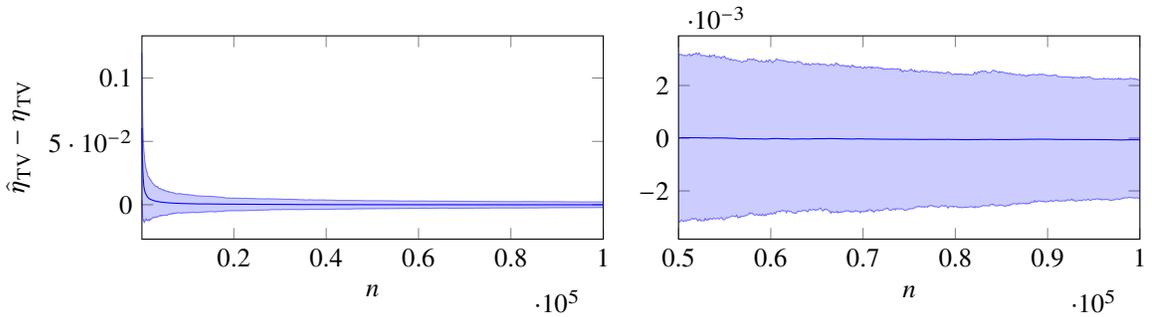

Figure 13: 5th percentile, mean, and 95th percentile of the difference of estimated and expected miscalibration of the uncalibrated model ($\beta_0 = \beta_1 = 1$) w.r.t. the total variation distance and 100 equally-sized bins (1000 series of random data).

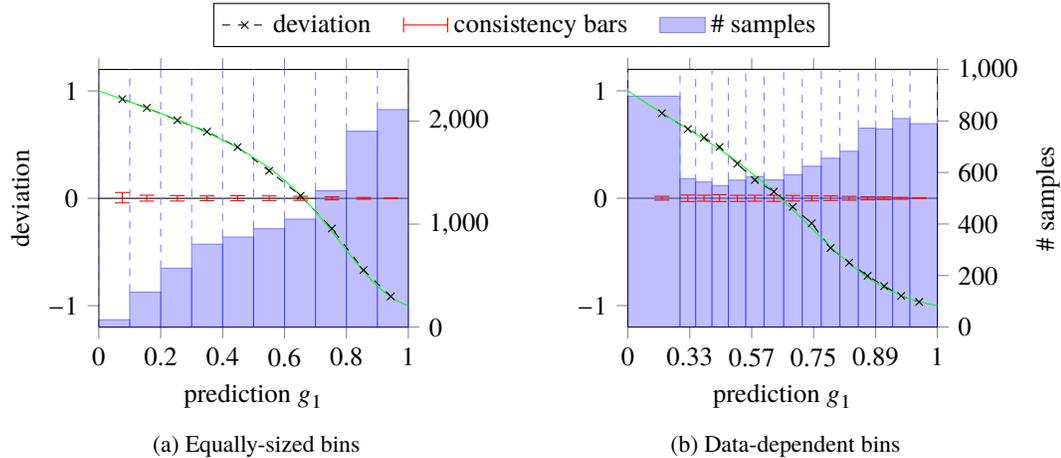

(a) Equally-sized bins

(b) Data-dependent bins

Figure 14: Reliability diagrams for the uncalibrated model ($\beta_0 = \beta_1 = 1$) w.r.t. the total variation distance on a randomly generated test set (10000 inputs). Dots indicate the deviation of the outcome distribution from the predictions in each bin. Blue bars show the distribution of predictions. Red bars visualize the 5th and 95th percentiles of the deviation in 1000 consistency resamples. The green curve shows the true analytical deviation.

## C.2 Neural network models

### C.2.1 DenseNet on CIFAR-10

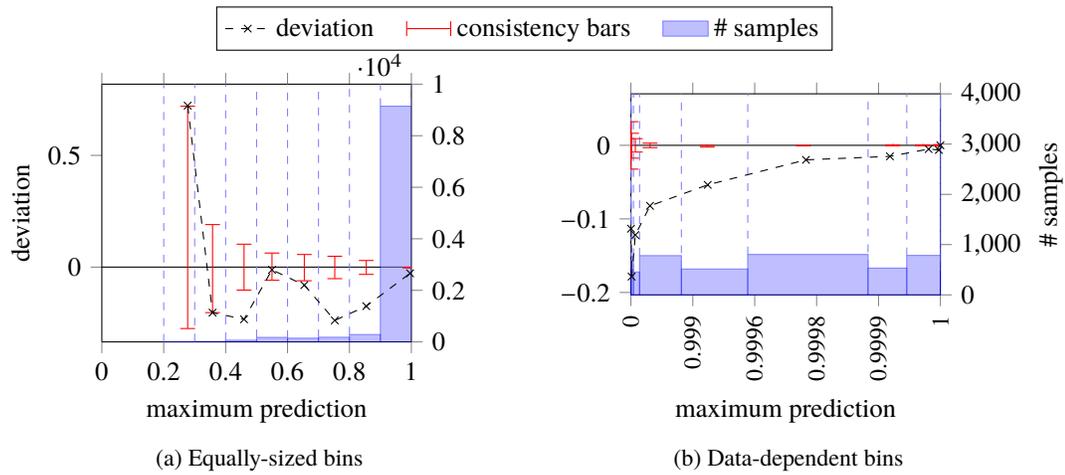

(a) Equally-sized bins

(b) Data-dependent bins

Figure 15: Reliability diagrams for the maximum predictions of DenseNet on the CIFAR-10 test set w.r.t. the total variation distance. Crosses indicate the deviation of the outcome distribution from the predictions in each bin. Blue bars show the distribution of predictions. Red bars visualize the 5th and 95th percentiles of the deviation in 1000 consistency resamples.

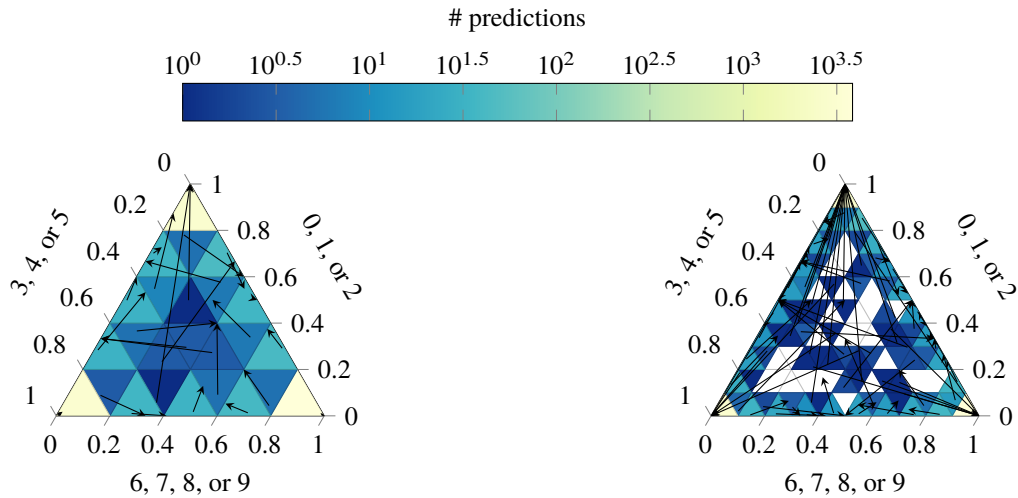

Figure 16: Two-dimensional reliability diagrams for DenseNet on the CIFAR-10 test set with 25 and 100 bins of equals. The predictions are grouped into three groups $\{0, 1, 2\}$, $\{3, 4, 5\}$, and $\{6, 7, 8, 9\}$ of the original classes. Arrows represent the deviation of the estimated calibration function value (arrow head) from the group prediction average (arrow tail) in a bin. The empirical distribution of predictions is visualized by color-coding the bins.

### C.2.2 ResNet on CIFAR-10

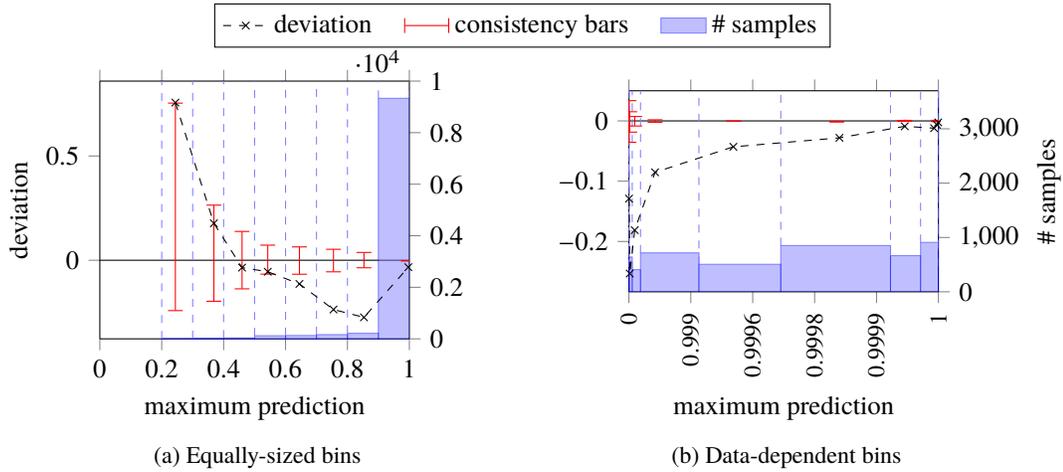

Figure 17: Reliability diagrams for the maximum predictions of ResNet on the CIFAR-10 test set w.r.t. the total variation distance. Crosses indicate the deviation of the outcome distribution from the predictions in each bin. Blue bars show the distribution of predictions. Red bars visualize the 5th and 95th percentiles of the deviation in 1000 consistency resamples.

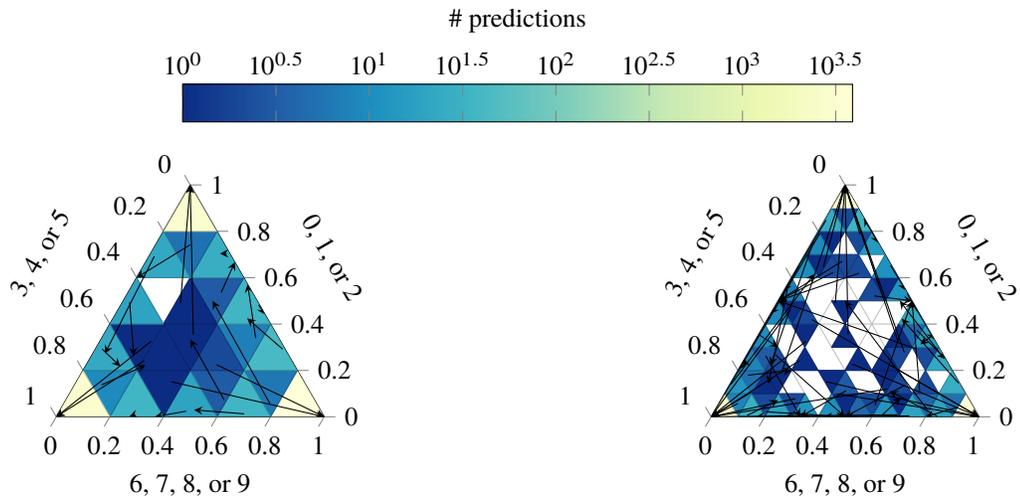

Figure 18: Two-dimensional reliability diagrams for ResNet on the CIFAR-10 test set with 25 and 100 bins of equal size. The predictions are grouped into three groups $\{0, 1, 2\}$, $\{3, 4, 5\}$, and $\{6, 7, 8, 9\}$ of the original classes. Arrows represent the deviation of the estimated calibration function value (arrow head) from the group prediction average (arrow tail) in a bin. The empirical distribution of predictions is visualized by color-coding the bins.

## D  Expected miscalibration estimates for neural networks

The results presented in Tables 3 to 5 show estimates of the expected miscalibration $\eta$ for DenseNet, ResNet, and LeNet models trained on CIFAR-10, using different binning schemes (see Appendix B), calibration lenses (see Example 2), and distance functions (see Section 3.2). Estimates with respect to the total variation distance for maximum predictions and equally-sized bins correspond to the expected miscalibration error used by Guo et al. (2017). For comparison, we also approximate estimates of the expected miscalibration under the consistency assumption of a perfectly calibrated model.

Standard deviations are estimated using bootstrapping. The accuracy of the investigated DenseNet, ResNet, and LeNet models is $0.933 \pm 0.002$, $0.934 \pm 0.002$, and $0.727 \pm 0.004$, respectively.

Table 3: Estimates of the expected miscalibration for DenseNet trained on CIFAR-10.

| Calibration lens | Distance $d$ | Equally-sized bins | | Data-dependent bins | |
|---|---|---|---|---|---|
| | | $\hat{\eta}_d$ | $\hat{\eta}_d^{\text{id}}$ | $\hat{\eta}_d$ | $\hat{\eta}_d^{\text{id}}$ |
| Canonical | Total variation | $0.059 \pm 0.002$ | $0.029 \pm 0.001$ | $0.041 \pm 0.002$ | $0.007 \pm 0.001$ |
| | Squared Euclidean | $0.072 \pm 0.003$ | $0.034 \pm 0.001$ | $0.046 \pm 0.003$ | $0.006 \pm 0.001$ |
| Maximum | Total variation | $0.038 \pm 0.002$ | $0.002 \pm 0.001$ | $0.038 \pm 0.002$ | $0.001 \pm 0.001$ |
| | Squared Euclidean | $0.054 \pm 0.003$ | $0.006 \pm 0.001$ | $0.053 \pm 0.003$ | $0.004 \pm 0.001$ |

Table 4: Estimates of the expected miscalibration for ResNet trained on CIFAR-10.

| Calibration lens | Distance $d$ | Equally-sized bins | | Data-dependent bins | |
|---|---|---|---|---|---|
| | | $\hat{\eta}_d$ | $\hat{\eta}_d^{\text{id}}$ | $\hat{\eta}_d$ | $\hat{\eta}_d^{\text{id}}$ |
| Canonical | Total variation | $0.059 \pm 0.002$ | $0.022 \pm 0.001$ | $0.042 \pm 0.002$ | $0.007 \pm 0.001$ |
| | Squared Euclidean | $0.071 \pm 0.003$ | $0.028 \pm 0.001$ | $0.047 \pm 0.003$ | $0.005 \pm 0.001$ |
| Maximum | Total variation | $0.043 \pm 0.002$ | $0.002 \pm 0.001$ | $0.043 \pm 0.002$ | $0.001 \pm 0.001$ |
| | Squared Euclidean | $0.061 \pm 0.003$ | $0.004 \pm 0.001$ | $0.061 \pm 0.003$ | $0.004 \pm 0.001$ |

Table 5: Estimates of the expected miscalibration for LeNet trained on CIFAR-10.

| Calibration lens | Distance $d$ | Equally-sized bins | | Data-dependent bins | |
|---|---|---|---|---|---|
| | | $\hat{\eta}_d$ | $\hat{\eta}_d^{\text{id}}$ | $\hat{\eta}_d$ | $\hat{\eta}_d^{\text{id}}$ |
| Canonical | Total variation | $0.219 \pm 0.003$ | $0.215 \pm 0.003$ | $0.027 \pm 0.003$ | $0.023 \pm 0.002$ |
| | Squared Euclidean | $0.243 \pm 0.004$ | $0.238 \pm 0.003$ | $0.024 \pm 0.003$ | $0.019 \pm 0.002$ |
| Maximum | Total variation | $0.007 \pm 0.003$ | $0.010 \pm 0.002$ | $0.009 \pm 0.003$ | $0.011 \pm 0.002$ |
| | Squared Euclidean | $0.010 \pm 0.004$ | $0.015 \pm 0.003$ | $0.013 \pm 0.004$ | $0.011 \pm 0.003$ |